# SpanRE: Entities and Overlapping Relations Extraction Based on Spans and Entity Attention

**Hao Zhang**
ZheJiang University / HangZhou, China

## Abstract

Extracting entities and relations is an essential task of information extraction. Triplets extracted from a sentence might overlap with each other. Previous methods either didn't address the overlapping issues or solved overlapping issues partially. We define a triplet as (subject, predicate, object) . To tackle triplet overlapping problems completely, firstly we extract candidate subjects with a standard span mechanism. Then we present a labeled span mechanism to extract the objects and relations simultaneously, we use the labeled span mechanism to generate labeled spans whose start and end positions indicate the objects, and whose labels correspond to relations of subject and objects. Besides, we design an entity attention mechanism to enhance the information fusion between subject and sentence during extracting objects and relations. We test our method on two public datasets, our method achieves the state-of-the-art performances on these two datasets.

## 1 Introduction

Triplets are the building blocks of large-scale and reusable knowledge bases which are beneficial for many natural language processing problems, such as pretrained language model (Zhang et al., 2019), question answering (Cui et al.; Yih et al., 2015) and recommendation model (Shen et al., 2019). Great efforts have been dedicated to the extraction of triplets from natural language text. The task of triplet extraction aims at identifying triplets like ( Albert Einstein, born_in, Germany) from text.

A sentence usually contains multiple triples. According to the overlapping situation of multiple triplets, the triplets can be divided into three categories: Normal, SingleEntiyOverlap (SEO) and EntityPairOverlap (EPO) (Zeng et al., 2018). Examples are shown in Figure 1. In a sentence, a triplet is Normal if the two entities of the triplet are not overlapping with other triplets; if a triplet has only one single overlapped entity and don't has overlapped entity pair, the triplet is SEO; a triplet is EPO if the triplet has the same entity pair with other triplets.

Most previous approaches decomposed the extraction task into two individual subtasks: entity recognition and relation classification, and the two subtasks are executed in a pipelined manner. However, the pipelined solution neglected relationship between entity recognition and relation classification, and suffered from error propagation (Li and Ji, 2014; Gupta et al., 2016). Recent studies mainly focused on extracting entities and relations jointly. Joint models (Yu and Lam, 2010; Ren et al., 2017) integrated the information of entity and relation, and have achieved a better performance. But most existing joint models need to carefully design features and heavily rely on other NLP toolkits.

In recent years, neural models have been applied in extracting entities and relations successfully, but it is not easy to build an elegant architectures to tackle triplet overlapping problems. Dai et al. (2019) presented a unified method to solve the joint extraction by tagging entity and relation labels simultaneously according to a query word position. However they can not solve the EPO issue, and their model built a labeling sequence for every word, which is very expensive to generate. Zeng et al. (2018) proposed an end-to-end model based on sequence-to-sequence learning with copy mechanism. Although it can solve all overlapping problems, their model can not extract multi-word entities. Takanobu et al. (2019) applied a hierarchical reinforcement learning framework to extract entities and relations, but they can not extract accurate triplets when a sentence has two or more triplets with the same relation.

In this paper, we present SpanRE, a novel joint extraction model based on span mechanism to solve all triplet overlapping issues, we also design an entity attention mechanism to fuse the information of entity and sentence. Wang and Jiang (2016) proposed Span mechanism, and used the Pointer Network (Vinyals et al., 2015) to predict the start and end position of an answer span in machine reading comprehension. Span-based model assigns a label for each token, tokens between two identical labels represent a span, so the spans can be used to detect entities. We define a triplet as (subject, predicate, object). Firstly, our model extracts all candidate subjects with standard span mechanism. Secondly, to extract the objects and relations corresponding to the candidate subject, we propose an entity attention mechanism to enhance the fusion of the candidate subject and sentence. We also present a novel labeled span mechanism which can extract objects and relations simultaneously. The key contributions of our work are as follows:

· We propose a novel joint model with span mechanisms and entity attention mechanism to extract triplets, where the entities and relations are extracted simultaneously.

· We propose a labeled span mechanism to solve all possible triplet overlapping cases including SingleEntiyOverlap and Entity-PairOverlap elegantly and accurately.

· We present an entity attention mechanism to fuse information of entity and sentence, the experiment demonstrates that our entity attention mechanism improves the result.

· We test our model on two public datasets. The experiment results show that our model achieves the state-of-the-art performance on the two datasets.

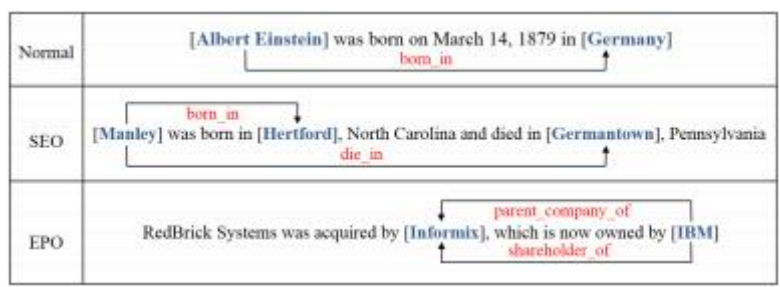

Figure 1: Examples of Normal, SingleEntityOverlap (SEO) and EntityPairOverlap (EPO). The entities are marked in blue, the relations are marked in red. Relations are unidirectional, so we use line with arrow to connect entities.

## 2    Related Work

For sentences with annotated entities, relation classification has been as a separate task to be tackled. Zeng et al. (2014); Xu et al. (2015a); Santos et al. (2015) took the task as multi-class classification problem. Zeng et al. (2014) exploited a convolutional neural network to extract lexical and sentence-level features used for relation classification. Santos et al. (2015) proposed a CR-CNN model to tackle relation classification task. Xu et al. (2015a) used a convolution neural network to learn more robust relation representations from the shortest dependency path, and improved the assignment of subjects and objects by negative sampling strategy. These models succeeded to identify the relation for entity pair, but they all needed annotated sentences.

For sentences without annotated entities, Nadeau and Sekine (2007); Chan and Roth (2011) separated entities and relations extraction into two subtasks and used pipelined manner to conduct the task, they neglected the relevance of the two subtasks. Recently, the joint model has made great success at entities and relations extraction. Primitive joint models depended on hand-craft features which were hard to construct. Li and Ji(2014) used structured perceptron with efficient beam-search and some global features to extract entities and relations jointly. Miwa and Sasaki (2014) proposed a table representation of entities and relations, they also incorporated global features, search orders and learning methods with inexact search on the table.

In recent studies, Zhang et al. (2017) proposed a new LSTM features and build a globally optimized neural model for end-to-end relation extraction. Miwa and Bansal (2016) proposed a neural model that shares parameters for entity extraction and relation classification. Most of above models can just extract Normal or SEO triplets, they never discussed the EPO issue. Some lastest models try to solve overlapping triplet extraction issues. Zeng et al. (2018) proposed an end-to-end model based on sequence-to-sequence learning with copy mechanism, their model can solve all triplet overlapping cases, but the model can not extract multi-word entities. Dai et al. (2019) designed a new tag schema to extract entities and relations simultaneously, they built a tag sequence for every word, their model got the corresponding entities and relations according to a query word. However they can only solve the SEO issue, and the model is very expensive to conduct. Takanobu et al. (2019) applied a hierarchical reinforcement learning framework to extract entities and relations. However their model can not handle the situation that a sentence has two or more triplets with same relation well. Besides, their model is very complex and expensive.

Our SpanRE model based on span mechanisms can solve all overlapping issues and is very cheap to conduct. Span-based model can tackle triplet overlapping problems and nested entities problem. Luan et al.(2018) introduced the annotated SciERC Dataset and used a span-based model for jointing coreference resolution, entity and relation extraction. A recent study (Luan et al., 2019) used the same span representation and added a graph propagation step to capture the interaction of spans. Different from above models, our model proposes a novel span structure which can extract objects and relations simultaneously.

# 3 Model

## 3.1 Model Overview

We denote an input sentence as $s = (w_1, w_2, ..., w_T)$, where $w_i$ with $i = 1, 2, .., T$ is the ith word of the sentence. Our model aims at extracting triplets from $s$. Triplets might overlap with each other. We present SpanRE, a novel joint extraction model based on span mechanisms and an entity attention mechanism to tackle this problem.

As illustrated in Figure 2, we first use the representation layer to generate feature vectors of the input sentence; secondly, we use a standard span mechanism to predict the start tag sequence **sub**start and the end tag sequence **sub**end of subject spans, and use the **sub**start and the **sub**end to extract subjects; then our entity attention mechanism fuses subject features and sentence features to predict the start tag sequence **obj–rel**start and the end tag sequence **obj–rel**end of object and relation spans with a novel labeled span mechanism, and we use the **obj–rel**start and the **obj–rel**end to extract objects and relations simultaneously.

## 3.2 Representation Layer

RNN-based models have achieved great success in NLP tasks, but RNN-based models are hard to be trained with full parallelization. As presented in Figure 2, we use the gated relation network (GRN) proposed by Chen et al. (2019) as the feature representation layer. The GRN model is based on CNN, and can be trained in parallel. In our model, the input layer is comprised of word-level features and character-level features. We initialize word-level features with GloVe[1] (Pennington et al., 2014) and they will be fine-tuned during training. Word-level feature of each word $w_i$ is represented as $s_i$.

The character-level features can solve the out-of-vocabulary problem (Rei et al., 2016; Ma and Hovy, 2016), the GRN model used a CNN to extract character-level features for each word and used a max-pooling to reduce the CNN results into a single embedding vector $\bar{u}_i$. Different from GRN, we use three different kernel sizes to capture the information in multi-scale, and perform a max-over-time pooling operation on the result of each kernel size:

$$\bar{c}^k_{i,j} = \text{conv}_k([c_{i,j\,-k/2}, ..., c_{i,j}, ..., c_{i,j+k/2}]) \quad (1)$$

$$u^k_i = \text{maxpooling}([\ \bar{c}^k_{i,0}, ..., \bar{c}^k_{i,j}, ..., \bar{c}^k_{i,L}]) \quad (2)$$

$$\bar{u}_i = [u^1_i; u^2_i; u^3_i] \quad (3)$$

where $k \in \{1, 2, 3\}$ is the kernel size of CNN layer. $c_{i,j}$ represents the embedding of jth character of ith word, L is the character length of ith word. The word feature vectors for a sentence are represented as $Z = \{z_1, z_2, ..., z_T\}$, where $z_i = [\bar{u}_i; s_i]$.

To capture local information of different scales, the GRN model adopted the context layer with multi-branch where each branch contained a CNN layer. Each branch extracted the local information $\bar{z}^k_i$ respectively, and then a max-pooling was used to select the strongest signals from all branches.

$$\bar{z}^k_i = \text{conv}_k([z_{i-k/2}, ..., z_i, ..., z_{i+k/2}]) \quad (4)$$

$$x_i = \text{maxpooling}(\text{tanh}\ ([\ \bar{z}^1_i, \bar{z}^3_i, \bar{z}^5_i])) \quad (5)$$

where $k \in \{1, 3, 5\}$ is the kernel size of CNN.

---

[1] http. - - nlp_ st_ nfora_ eau- projebts- glove-

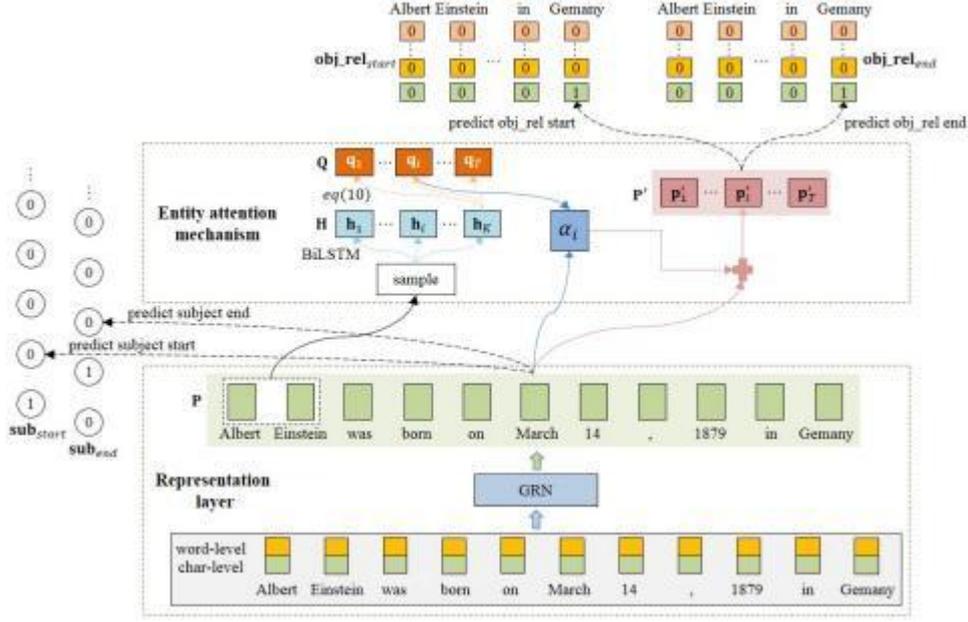

Figure 2: The architecture of SpanRE, this figure contains representation layer, entity attention mechanism, subject prediction and obj_rel prediction. The meanings of the annotations in the figure are consistent with the context. Best seen in color.

As we all know CNN cannot capture the long-term context information, to tackle this problem, the GRN employed a gated relation layer to enhance the conventional CNNs with global context information. Given the local context feature vectors $X = c_{x_1}, x_2, ..., x_T$, the gated relation layer firstly computed the relation score $r_{ij}$ between word $w_i$ and $w_j$, and then used the gating mechanism to compute feature vector $p_i$ of $i$th word:

$$r_{ij} = \sigma(W_{r \cdot t} [x_i; x_j] + b_{r \cdot t}) \quad (6)$$

$$p_i = \tanh(\frac{1}{T}\sum_{j=1}^{T} r_{ij} * x_j) \quad (7)$$

where $\sigma$ is the sigmoid function, $P = c_{p_1}, ..., p_T$ is the final feature vectors of input sentence.

### 3.3 Entity Attention Mechanism

We think that the information of a subject and sentence is essential to extract objects and relations corresponding the subject. So we present an entity attention mechanism to fuse the information of subject and sentence. The entity attention mechanism is shown in Figure 2. Firstly, we set K as the length of subject feature vectors, for each subject whose representation feature vectors are described as $B = c_{p_s}, ..., p_e$, where s is the start position of the subject and e is the end position of the subject, we sample K word feature vectors from $B$ to compose the subject feature vectors $B$. In view of K << T, to get more semantic information, we use a BiLSTM to encode $B$ as $H = c_{h_1}, h_2, ..., h_K$, we extend last step hidden state $h_K$ to the same length as the sentence. In our experience, the positions between subject and other words contain rich information, so our model adds the relative position embedding to the extended hidden state, the final subject feature vectors are represented as $Q = c_{q_1}, q_2, ..., q_T$:

$$\overline{B} = \text{sample}(B, K) \quad (8)$$

$$H = \text{BiLSTM}(\overline{B}) \quad (9)$$

$$Q = \text{repeat}(h_K, T) + R \quad (10)$$

where sample($B$, K) means sampling K feature vectors from $B$, repeat($h_K$, T) means repeating $h_K$ for T times, $R$ representes relative position embedding.

Given the sentence final feature vectors $P = c_{p_1}, p_2, ..., p_T$ and subject feature vectors $Q = c_{q_1}, q_2, ..., q_T$. We combine $Q$ and $P$ via multiplication attention as follows:

$$s_{ij} = S(q_i, p_j) \quad (11)$$

$$\alpha_{ij} = \frac{\exp(s_{ij})}{\sum_k \exp(s_{ik})} \quad (12)$$

$$p_i^{\setminus} = \sum_j \alpha_{ij} p_j \quad (13)$$

where $S(\mathbf{x}, \mathbf{y}) = \mathbf{x}^T \mathbf{U}^T \mathbf{V} \mathbf{y}$, $\mathbf{U}, \mathbf{V} \in R_k^{\times d}$, $k$ is the attention hidden size, d is the feature size. Fused feature vectors for extracting objects and relations are represented as $\mathbf{P}^\backslash = (\mathbf{p}_1^\backslash, \mathbf{p}_2^\backslash, ..., \mathbf{p}_i^\backslash)$.

$$G = \frac{add\text{-}word\text{-}exp \quad del\text{-}word\text{-}exp}{word \quad epx}$$

### 3.4 Labeled Span Mechanism and Triplets Extraction

In our SpanRE model, we propose two different span structures to extract subject and obj_rel (object and relation) respectively. We use the standard span mechanism to predict the start and end positions of subject spans. More specifically, we use two MLP layers to predict the start and end positions of subject spans. Different from extracting subjects, we design a novel labeled span mechanism to extract objects and relations, we use two fully connected layers to generate the start and end tag sequences of obj_rel. Figure 3 illustrates the details of the two span structures used in our model.

As described in Figure 3, to extract subjects, we assign a start sequence $\mathbf{sub}_{start}$ = $(st_1, st_2, ..., st_T)$ and an end sequence $\mathbf{sub}_{end}$ = $(ed_1, ed_2, ..., ed_T)$, where $st_i$ and $ed_i$ are 0 or 1. If $st_i$ is 1, it means the ith word is the start of a subject span; if $ed_i$ is 1, it means the ith word is the end of a subject span. We use the subject span to extract a candidate subject. A novel labeled span mechanism is designed for extracting object and relation simultaneously. In the labeled span mechanism, we assign a start tag sequence $\mathbf{obj}\text{-}\mathbf{rel}_{start}$ = $(\mathbf{st}_1^\backslash, \mathbf{st}_2^\backslash..., \mathbf{st}_T^\backslash)$ and an end tag sequence $\mathbf{obj}\text{-}\mathbf{rel}_{end}$ = $(\mathbf{ed}_1^\backslash, \mathbf{ed}_2^\backslash ..., \mathbf{ed}_T^\backslash)$, where $\mathbf{st}_i^\backslash$ and $\mathbf{ed}_i^\backslash$ are tag sequences whose dimensions are the number of relations, each element of $\mathbf{st}_i^\backslash$ and $\mathbf{ed}_i^\backslash$ are 0 or 1. If the jth element of $\mathbf{st}_i^\backslash$ is 1, we take the ith word as the first word of an object, and the object corresponds to jth relation; if the jth element of $\mathbf{ed}_i^\backslash$ is 1, we take the ith word as the last word of an object, and the object corresponds to jth relation. $\mathbf{st}_i^\backslash$ and $\mathbf{ed}_i^\backslash$ might have more than one 1, so an object can corresponds to different relations. Our labeled span mechanism can naturally solve triplet overlapping problems completely.

For example, as illustrated in Figure 3, start tag of word "IBM" is 1 in $\mathbf{sub}_{start}$, end tag of word "IBM" is 1 in $\mathbf{sub}_{end}$, so we extract the word "IBM" as a candidate subject. Then we use entity attention mechanism to combine information of candidate subject "IBM" and sentence to generate obj_rel start tag sequences and obj_rel end tag sequences. There are two 1 values in obj_rel start tag sequence of word "Informix", and the positions of 1 correspond to relations "parent_company_of" and "shareholders_of" respectively, the end tag sequence of word "Informix" correspond to relations "parent_company_of" and "shareholders_of" too. We find the same relation from the $\mathbf{obj}\text{-}\mathbf{rel}_{start}$ and $\mathbf{obj}\text{-}\mathbf{rel}_{end}$ to extract objects and relations. We extract two triplets (IBM, parent_company_of, Informix) and (IBM, shareholders_of, Informix). These two triplets are EPO, the example demonstrates the ability that our model can address all triplet overlapping cases.

### 3.5 Training

Both subject extraction and obj_rel extraction use 0-1 value tag sequences as target sequences, so the loss is calculated by binary cross entropy. The subject extraction loss and the obj_rel extraction loss both consist of start tag sequences loss and end tag sequences loss. Our model uses the sum of the subject extraction loss and the obj_rel extraction loss as the optimization objective. The objective function can be defined as:

$$e = e_{\mathbf{sub}_{start}} + e_{\mathbf{sub}_{end}} \\ + e_{\mathbf{obj}\_\mathbf{rel}_{start}} + e_{\mathbf{obj}\_\mathbf{rel}_{end}} \quad (14)$$

where $e_{\mathbf{sub}_{start}}$ is the binary cross entropy loss of subject start sequence, $e_{\mathbf{sub}_{end}}$ is the binary cross entropy loss of subject end sequence, $e_{\mathbf{obj}\_\mathbf{rel}_{start}}$ is the binary cross entropy loss of obj_rel start sequences, $e_{\mathbf{obj}\_\mathbf{rel}_{end}}$ is the binary cross entropy loss of obj_rel end sequences.

## 4 Experiments

### 4.1 Dataset

We evaluate the effectiveness of our model on two widely used public datasets: NYT[2] (Riedel et al., 2010) and WebNLG[3] (Gardent et al., 2017).

NYT (Riedel et al., 2010) is a news corpus sampled from about 294k New York Times news articles. We use the dataset published by Ren et al. (2017), and filter out the triplets whose relations are "None". We randomly sample 5000 sentences from the training data as testing data, and 5000

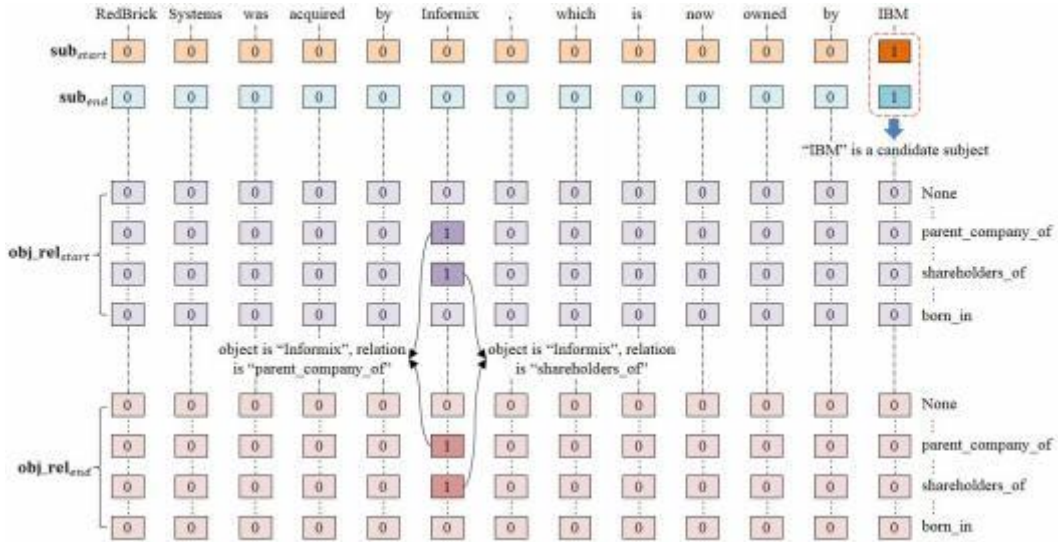

Figure 3: An example of our span mechanisms. For this sentence, firstly we use **sub**start and **sub**end to extract candidate subject "IBM", secondly we combine the candidate subject "IBM" and sentence to generate **obj–rel**start and **obj–rel**end, we extract two triplets (IBM, parent–company–of, Informix), (IBM, shareholders_of, Informix) from this sentence.

| Dataset | NYT | WebNLG |
|---|---|---|
| # Relations | 24 | 159 |
| # Training sentences | 56336 | 10580 |
| # Training Normal | 36439 | 5043 |
| # Training SEO | 33283 | 18259 |
| # Training EPO | 11173 | 62 |
| # Testing sentences | 5000 | 1479 |
| # Testing Normal | 3291 | 624 |
| # Testing SEO | 2901 | 2402 |
| # Testing EPO | 909 | 10 |

Table 1: Statistics of the datasets

sentences as validation data, the rest sentences are used to train our model.

WebNLG (Gardent et al., 2017) was originally published for natural language generation task. We use the dataset published by Zeng et al. (2018). The original WebNLG dataset includes 6940 entries in training file and 872 entries in testing file, every entry contains a group of triplets and several standard sentences (written by human). We combine the triplets with every standard sentence to process data. In addition, we take the data coming from testing file as testing data, and we randomly sample 1000 sentences from data coming from training file as validation data, the rest data from training file are as training data.

The statistics of the two processed datasets are presented in Table 1.

## 4.2 Baselines

We compare our model with several previous state-of-the-art methods on NYT and WebNLG, we get the codes from open source and conduct them by ourselves.

**LSTM-LSTM-Bias** (Zheng et al., 2017) took the joint extraction tasks as sequence labeling problem based on a novel tagging schema. The method can extract entities and relations in a step, however it can not solve the overlapping problems.

**CopyR** (Zeng et al., 2018) used seq2seq framework with a copy mechanism to extract triplets, and can address overlapping problems, but CopyR can not extract multi-word entities.

**PA-LSTM-CRF** (Dai et al., 2019) designed a tagging scheme to generate n tag sequences for a n-words sentence, the model also used a position-attention mechanism to produce different sentence representations and extracted the entities and relations simultaneously. The model can not solve EPO issue, and was expensive to conduct.

**HRL** (Takanobu et al., 2019) used hierarchical reinforcement learning framework to enhance interaction of entities and relations. HRL model detected the relations firstly, and then extracted entities corresponding to each relation. When there are two or more triplets with same relation in a sentence, HRL would extract inaccurate entity combination, besides HRL was very complex and expensive to conduct.

| Model | NYT | | | | WebNLG | | | |
|---|---|---|---|---|---|---|---|---|
| | *Precise* | *Recall* | *F1* | cvg_time | *Precise* | *Recall* | *F1* | cvg_time |
| LSTM-LSTM-Bias (Zheng et al., 2017) | 0.664 | 0.536 | 0.593 | 0.73h | - | - | - | - |
| CopyR (Zeng et al., 2018) | 0.596 | 0.536 | 0.564 | 2.14h | 0.186 | 0.142 | 0.161 | 0.71h |
| PA-LSTM-CRF (Dai et al., 2019) | 0.702 | 0.616 | 0.656 | 2.67h | - | - | - | - |
| HRL (Takanobu et al., 2019) | 0.665 | **0.743** | 0.702 | 19.2h | 0.66 | **0.68** | 0.669 | 6.68h |
| **SpanRE** | **0.797** | 0.715 | **0.754** | **0.69h** | **0.775** | 0.609 | **0.682** | **0.56h** |

Table 2: Main result of triplets extraction. The cvg_time is the short of convergence time, we execute all models with the default parameters of the codes under the same operating conditions. The unit of cvg_time is hour. SpanRE denotes our model.

### 4.3 Evaluation Metrics

Our model extracts entities and relations jointly, and we don't focus on entity type, so we evaluate our model by the performance of triplets. A triplet is taken as correct when its two entities and corresponding relation are all correct, where the entity is correct when the start position and end position of the entity are all correct. We adopt standard Precision, Recall and F1 score to evaluate the results. We exclude all triplets with relation of "None".

### 4.4 Parameter Settings

We initialize word-level features with GloVe (Pennington et al., 2014), the dimension of word-level feature is 300, and initialize the character-level features with the dimension of 300 randomly. We regularize our network using dropout on word features layer and the dropout ratio is 0.25. Other parameters of representation layer are same with GRN (Chen et al., 2019). In entity attention mechanism, we use 256 as the attention hidden size, and set subject sampling length as 4. The optimizer used in our model is Adam, the start learning rate is 1e-4. Besides, we use the warm up learning, we find that the warm up learning can reduce overfit and accelerate model convergence.

### 4.5 Main Results

The main results of different methods are shown in Table 2. Due to the data in WebNLG is lack of entity label, so we test the WebNLG only on CopyR, HRL and our SpanRE. Observing the results we can get that our SpanRE model achieves the state-of-the-art performances on the two public datasets in *Precise* score and *F1* score. On NYT dataset, comparing with previous best results, we achieve a 5.2% improvement in *F1* score, a 13.2% improvement in *Precise* score. On WebNLG dataset, comparing with previous best results, we improve 11.5% in *Precise* score and 1.3% in *F1* score. Under the same operating conditions, the convergence time indicates that our model is the fastest. The experiment show that the SpanRE model we proposed is effective. Besides, from Table 2, we can find that the performance of LSTM-LSTM-Bias (Zheng et al., 2017) is worse than other methods, the main reason is that LSTM-LSTM-Bias model can not extract overlapping triplets. CopyR (Zeng et al., 2018) model can not extract multi-word entities, so its results underperform. PA-LSTM-CRF (Dai et al., 2019) cannot solve EPO issue, and HRL (Takanobu et al., 2019) will extract inaccurate entity combination, our model can address all these problems, so we outperform. Our model achieves a better result on NYT dataset than WebNLG dataset, we believe the low proportion of EPO triplets on WebNLG lead to this situation.

We also study the performance of our model in different length of entity on NYT dataset. The experiment results are shown at Figure 4. With the length of entity increasing, the performances of all methods descend, but our model is more smooth and always keeps a high level performance. PA-LSTM-CRF (Dai et al., 2019) and HRL (Takanobu et al., 2019) perform very badly when the length of entity is more than 2.

### 4.6 Overlapping Triplet Extraction

According to Table 1, there are 3291 Normal triplets, 2901 SEO triplets and 909 EPO triplets in NYT testing data. There are 624 Normal triplets, 2402 SEO triplets and 10 EPO triplets in WebNLG testing data. The number of EPO triplets in WebNLG dataset is too little, so we just test the overlapping triplet extraction on NYT dataset. Because LSTM-LSTM-Bias (Zheng et al., 2017) can not solve overlapping issues, we test overlapping triplet extraction only comparing with CopyR (Zeng et al., 2018), PA-LSTM-CRF (Dai et al., 2019) and HRL (Takanobu et al., 2019). The results of overlapping triplet extraction experiment are shown in Table 3.

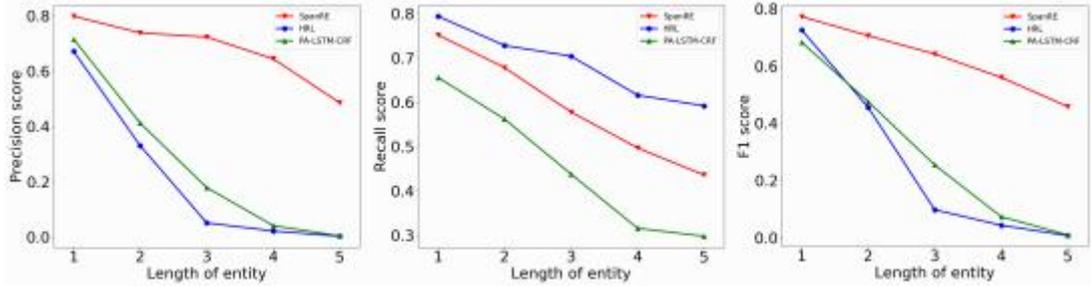

Figure 4: Metrics in different entity length. The entities whose length is more than 5 are very few, so we only show results with entity length less than 6.

| Model | NYT | | |
|---|---|---|---|
| | *Precise* | *Recall* | *F1* |
| CopyR (Zeng et al., 2018) | 0.355 | 0.489 | 0.411 |
| PA-LSTM-CRF (Dai et al., 2019) | 0.573 | 0.654 | 0.611 |
| HRL (Takanobu et al., 2019) | 0.583 | **0.755** | 0.658 |
| **SpanRE** | **0.619** | 0.733 | **0.672** |

Table 3: Results of overlapping triplets extraction experiment

| Model | NYT | | |
|---|---|---|---|
| | *Precise* | *Recall* | *F1* |
| **SpanRE** | **0.797** | **0.715** | **0.754** |
| - mulit-scale CNNs | 0.763 | 0.691 | 0.725 |
| - entity attention | 0.703 | 0.669 | 0.685 |

Table 4: Result of ablation experiment

Our model outperforms other methods. We get a 3.6% improvement in *Precise* score and a 1.4% improvement in *F1* score compared with the best previous model. This result mainly relies on our model can solve all overlapping problems, but other methods can not. It demonstrates the effectiveness of our method on extraction of overlapping triplets.

### 4.7 Ablation Study

In this section, we study the impact of components in our model. We just experiment on NYT dataset. Three experiments are conducted. 1) The base SpanRE is conducted just as Table 2. 2) We use original GRN (Chen et al., 2019) as representation layer, we extract the character-level features just using a CNN rather than using multi-scale CNNs. 3) We delete the entity attention mechanism in SpanRE, and concatenate **Q** (computed by eq(10)) and sentence representation **P** (computed by eq(7)) as the last representation layer. The experiment results are shown in Table 4.

Observing Table 4, we get that all the components of our model play important roles. Without using multi-scale CNNs to extract character-level features, the performance descends. It shows the multi-scale CNNs can get more information from characters. Comparing SpanRE and SpanRE without entity attention mechanism, we find that our entity attention mechanism is crucial for our SpanRE model. The evaluation metrics are bad without entity attention mechanism, which indicates that capturing the relationship between sentence and entity plays an important role for triplet extraction.

## 5 Conclusion and Future Work

In this paper, we present a novel joint model based on span mechanisms and an entity attention mechanism for triplets extraction. Our model can solve all triplet overlapping issues completely. To capture the interaction of entity and sentence, we design an entity attention mechanism, our entity attention mechanism is crucial for extracting triplets. We test our model on two public datasets, experiment results show that our model achieves the state-of-the-art performance. The experiment also shows that components of our model are all indispensable.

In the future, we will study the applications of our model in entity linking and entity disambiguation.